\documentclass[10pt,twocolumn,letterpaper]{article}

\usepackage{wacv}              

\usepackage{graphicx}
\usepackage{amsmath}
\usepackage{amssymb}
\usepackage{booktabs}

\usepackage[pagebackref,breaklinks,colorlinks]{hyperref}

\usepackage[capitalize]{cleveref}
\crefname{section}{Sec.}{Secs.}
\Crefname{section}{Section}{Sections}
\Crefname{table}{Table}{Tables}
\crefname{table}{Tab.}{Tabs.}

\def\wacvPaperID{916} 
\def\confName{WACV}
\def\confYear{2025}

\usepackage{xr}

\usepackage{color}

\usepackage[export]{adjustbox}
\usepackage{array}
\usepackage{arydshln}
\usepackage{booktabs}
\usepackage{colortbl}
\usepackage{float,wrapfig}
\usepackage{hhline}
\usepackage{multirow}
\usepackage{tabularx}
\usepackage{newfloat}

\usepackage{bm}
\usepackage{nicefrac}
\usepackage{microtype}
\usepackage{xfrac}

\usepackage{changepage}
\usepackage{extramarks}
\usepackage{fancyhdr}
\usepackage{setspace}
\usepackage{soul}
\usepackage{xspace}

\usepackage{etoolbox,siunitx}
\usepackage{afterpage}
\usepackage{subcaption}

\usepackage{calc}
\usepackage[title]{appendix}

\usepackage{extarrows}

\usepackage{lipsum}  
\usepackage{pifont}
\usepackage{enumitem}
\usepackage[ruled]{algorithm2e} %

\usepackage{physics}
\usepackage{makecell}
\usepackage{cleveref}
\usepackage{adjustbox}
\usepackage{enumitem}
\usepackage{lipsum} 

\newcommand{\ignorethis}[1]{}

\makeatletter
\DeclareRobustCommand\onedot{\futurelet\@let@token\@onedot}
\def\@onedot{\ifx\@let@token.\else.\null\fi\xspace}

\makeatother

\newcommand*{\rom}[1]{\expandafter\romannumeral #1}

\definecolor{mydarkblue}{rgb}{0,0.08,1}
\definecolor{mydarkgreen}{rgb}{0.02,0.6,0.02}
\definecolor{mydarkred}{rgb}{0.8,0.02,0.02}
\definecolor{mydarkorange}{rgb}{0.40,0.2,0.02}
\definecolor{mypurple}{RGB}{111,0,255}
\definecolor{myred}{rgb}{1.0,0.0,0.0}
\definecolor{mygold}{rgb}{0.75,0.6,0.12}
\definecolor{myblue}{rgb}{0,0.2,0.8}
\definecolor{mydarkgray}{rgb}{0.66,0.66,0.66}

\newif\ifcolor
\colortrue

\newif\ifdraft
\draftfalse

\ifdraft
    \newcommand{\kac}[1]{{\color{magenta}\textbf{Kfir:} #1}}
    \newcommand{\ync}[1]{{\color{blue}\textbf{Yotam:} #1}}
    \newcommand{\dcc}[1]{{\color{red}\textbf{Danny:} #1}}
    \newcommand{\ygc}[1]{{\color{cyan}\textbf{Yossi:} #1}}
    \newcommand{\imc}[1]{{\color{green}\textbf{Inbar:} #1}}
    \newcommand{\qhc}[1]{{\color{teal}\textbf{Charles:} #1}}
    \newcommand{\myc}[1]{{\color{teal}\textbf{Michal:} #1}}
    \newcommand{\olc}[1]{{\color{violet}\textbf{Orly:} #1}}
    \newcommand{\ypc}[1]{{\color{red}\textbf{Yael:} #1}}

    \newcommand{\nuke}[1]{{\color{red}#1}} %
    \newcommand{\move}[1]{{\color{orange}#1}} %
    
\else
    \newcommand{\kac}[1]{}
    \newcommand{\ync}[1]{}
    \newcommand{\dcc}[1]{}
    \newcommand{\ygc}[1]{}
    \newcommand{\imc}[1]{}
    \newcommand{\qhc}[1]{}
    \newcommand{\olc}[1]{}
    \newcommand{\ypc}[1]{}
    \newcommand{\myc}[1]{}
    \newcommand{\nuke}[1]{} %
    \newcommand{\move}[1]{} %

\fi

\newif\ifcamera
\camerafalse

\ifcamera
    \newcommand{\camera}[1]{#1}
\else
    \newcommand{\camera}[1]{}
\fi

\newcommand{\vect}[1]{\boldsymbol{\mathbf{#1}}}

\newlist{todolist}{itemize}{2}
\setlist[todolist]{label=$\square$}

\usepackage{titletoc}
\usepackage{multirow}

\SetAlFnt{\small}
\SetAlCapFnt{\small}
\SetAlCapNameFnt{\small}
\SetAlCapHSkip{0pt}

\newcommand{\notes}[1] {\textcolor{black}{#1}}
\newcommand{\change}[1] {\textcolor{black}{#1}}
\newcommand{\NFL}{NeRFFaceLighting}

\begin{document}

\title{Analyzing and Improving the Skin Tone Consistency and Bias in Implicit 3D Relightable Face Generators}

\author{Libing Zeng\\
Texas A\&M University\\
College Station, USA\\
{\tt\small libingzeng@tamu.edu}
\and
Nima Khademi Kalantari\\
Texas A\&M University\\
College Station, USA\\
{\tt\small nimak@tamu.edu}
}
\maketitle

\begin{abstract}

With the advances in generative adversarial networks (GANs) and neural rendering, 3D relightable face generation has received significant attention. Among the existing methods, a particularly successful technique uses an implicit lighting representation and generates relit images through the product of synthesized albedo and light-dependent shading images. While this approach produces high-quality results with intricate shading details, it often has difficulty producing relit images with consistent skin tones, particularly when the lighting condition is extracted from images of individuals with dark skin. Additionally, this technique is biased towards producing albedo images with lighter skin tones. Our main observation is that this problem is rooted in the biased spherical harmonics (SH) coefficients, used during training. 
\change{Following this observation, we conduct an analysis and demonstrate that the bias appears not only in band 0 (DC term), but also in the other bands of the estimated SH coefficients. We then propose a simple, but effective, strategy to mitigate the problem. Specifically, we normalize the SH coefficients by their DC term to eliminate the inherent magnitude bias, while statistically align the coefficients in the other bands to alleviate the directional bias. We also propose a scaling strategy to match the distribution of illumination magnitude in the generated images with the training data.}
Through extensive experiments, we demonstrate the effectiveness of our solution in increasing the skin tone consistency and mitigating bias.

\vspace{-0.2in}
\end{abstract}

\section{Introduction}
\label{sec:intro}

In recent years, there has been a growing interest in learning 3D generative models of faces from 2D images~\cite{Niemeyer2020GIRAFFE, piGAN2021, zhou2021CIPS3D, EG3D2021Chan, deng2022gram, xiang2023gramhd, gu2021stylenerf} through a combination of generative adversarial networks (GAN)~\cite{gan2014NIPS} and neural radiance fields~\cite{mildenhall2020nerf}. However, utilizing such 3D digital humans in various applications, such as virtual/augmented reality and gaming, necessitates full control over different image formation factors such as geometry, appearance, and lighting. 

Several approaches~\cite{pan2021shadegan, tan2022volux, FaceLit2023CVPR, Lumigan2023Deng} attempt to address this entanglement problem by incorporating analytical lighting models, such as Phong, into the 3D face generators. Specifically, the generator estimates the shading components, such as diffuse and specular, which are then used along with the lighting information in a shading model to produce the final images. Because of using analytical models, these approaches can produce consistent relit images that match the input illumination. However, due to computational costs, these techniques use simple shading models that ignore realistic effects such as subsurface scattering and interreflection. As a result, they often produce images that do not have the intricate shading details of real faces.

\begin{figure}
    \centering
    \includegraphics[width=1.0\linewidth]{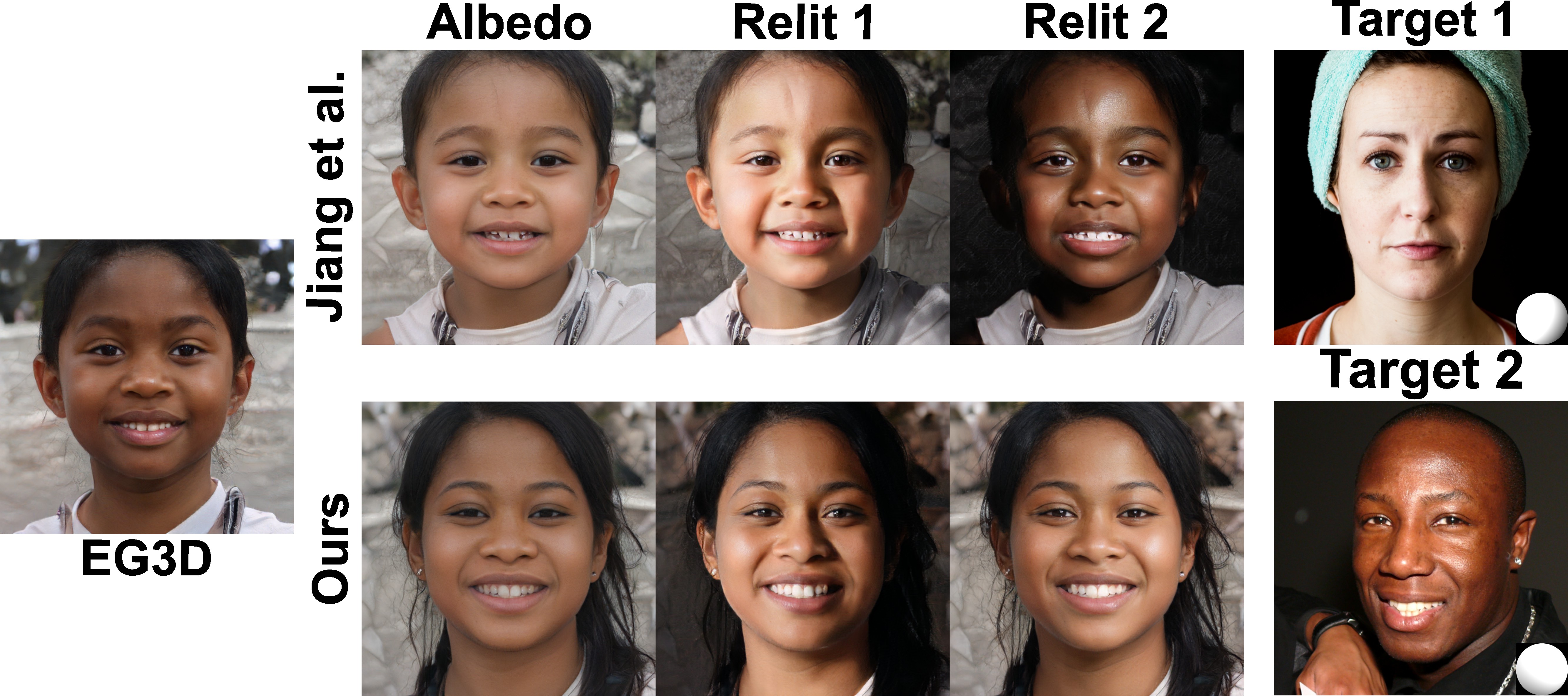} 
    \vspace{-0.25in}
    \caption{On the top, we show two relit images produced by \NFL{} (Jiang et al.)~\cite{NeRFFaceLighting2023Jiang}, using the lighting extracted from images of individuals with fair and dark skin tones (shown on the right). As seen, \NFL{} produces relit images with inconsistent skin tones. Additionally, when distilling the EG3D triplane, \NFL{} tends to produce albedo maps that are biased towards lighter skin colors. Our method mitigates this bias and improves the consistency of the skin tone in relit images. Note that even though we use the same latent vector to generate the results with EG3D, \NFL{}, and ours, there are variation in the images as the backbone EG3D network is fine-tuned separately in \NFL{} and ours.
    }
    \label{fig:teaser}
\vspace{-0.2in}
\end{figure}

The state-of-the-art method of Jiang et al.~\cite{NeRFFaceLighting2023Jiang} (\NFL{}) tackles this issue through an implicit lighting representation. The key idea is to distill the original single triplane of EG3D~\cite{EG3D2021Chan} into shading and albedo triplanes. The shading triplane is estimated through an adapter network that is built on top of the EG3D generator and additionally takes the lighting information in form of spherical harmonics (SH)~\cite{ravi2001sh} as the input. Training is performed using the original EG3D discriminator, along with another discriminator that is additionally conditioned on the SH coefficients. The extra conditional discriminator forces the generator to produce relit images that are consistent with the lighting. Due to the flexibility of the implicit lighting model, \NFL{} produces 3D portraits with more realistic lighting compared to methods using explicit lighting models.



Despite producing impressive results, \NFL{} suffers from two major drawbacks. First, this technique is not able to produce relit images with consistent skin tones. Specifically, when the target lighting is obtained from images with dark skin, \NFL{} produces portraits with a dark skin tone, regardless of the albedo's skin color. As shown in Fig.~\ref{fig:teaser}, the relit image corresponding to the target lighting from a dark skin individual (Target 2), has a significantly darker skin color than the albedo. Second, \NFL{} distills EG3D into an albedo representation that is biased towards lighter skin tones (see Figs.~\ref{fig:teaser}~and~\ref{fig:motivation_separation}).


Our key contribution is to conduct an analysis to pinpoint the root causes of these issues, and present a simple, yet effective, strategy to mitigate the problems. We observe that the two drawbacks are connected and the core issue lies in the bias of existing lighting estimation methods~\cite{sfsnetSengupta18}, which are used to generate the SH coefficients for training images. These techniques tend to estimate SH coefficients that exhibit correlation with skin color. The discriminator observes the correlation in real data and forces the generator to produce relit images with dependency between the skin color and lighting. Because of using implicit lighting representation, the generator in \NFL{} has the flexibility to change the skin color through shading. Furthermore, since the generator can produce individuals with dark skin color through shading, it opts for synthesizing albedo images with lighter skin tone.

Following this observation, we conduct an analysis and demonstrate that the bias manifests itself \change{not only in band 0 (DC term) of the estimated SH coefficients, but also in the other bands}. Specifically, estimated SH coefficients of individuals with darker skin have smaller DC term, indicating that existing light estimation methods are biased towards albedos with lighter skin and compensate the difference by estimating dimmer lighting. \change{Moreover, the coefficients in other bands of images with dark skin form a distinct cluster, indicating directional bias in the estimated coefficients.} 




\change{We propose to address the magnitude and directional biases by utilizing normalized and statistically-aligned SH coefficients as input for both the generator and discriminator during training.}
Specifically, we normalize the coefficients by the DC term to ensure a constant average for all lightings, thereby alleviating the bias in illumination magnitude.
\change{To address the bias in the other bands, we make a key observation that the illumination in the training data is independent of the skin tone. As such, the illumination of the dark and non-dark skin tones should be statistically similar. Based on this observation, we propose to statistically align the SH coefficients of the individuals with dark skin tones to the ones corresponding to images with non-dark skin colors.}

We perform the training using these normalized and statistically-aligned SH coefficients. One potential problem is that the generator takes normalized SH coefficients, and thus should produce relit images with constant illumination magnitude. However, the illumination magnitude of the training images vary significantly. To address this issue, we utilize the linearity of light and propose to scale the generated relit images, thereby scaling their lighting, to match the distribution of the training data.  

Once trained, our generator can produce relit images consistent with the direction of illumination, although it is limited to handling lighting with a constant magnitude. To accommodate illuminations with arbitrary magnitudes, we use our generator to produce results under normalized lighting and then directly scale the relit images to the appropriate magnitude. We demonstrate that these modifications significantly improve the consistency of skin tone in relit images under varying lighting conditions and alleviate the albedo generation bias toward lighter skin tones (see Fig.~\ref{fig:teaser}). Moreover, we demonstrate that our approach can be used to improve the skin tone consistency of other relighting approaches like, DiFaReli~\cite{ponglertnapakorn2023difareli}. \emph{We will release the source code upon publication.}

\section{Related Work}
\label{sec:relatedwork}

In this section, we discuss the closely related work on 3D generative networks, 3D relightable face generators, and \change{portrait relighting} approaches. We also provide a brief review of the algorithms that address bias in light/albedo estimation.



\subsection{3D Generative Networks}

Generative Adversarial Networks (GAN)~\cite{gan2014NIPS}, in particular StyleGAN~\cite{karras2019style, karras2020analyzing, karras2021alias}, are capable of producing results that are virtually indistinguishable from real images. A large number of techniques~\cite{Niemeyer2020GIRAFFE, piGAN2021, zhou2021CIPS3D, EG3D2021Chan, deng2022gram, xiang2023gramhd, gu2021stylenerf} combine GAN with neural radiance field (NeRF)~\cite{mildenhall2020nerf} to synthesize 3D consistent high-quality images. In particular, EG3D~\cite{EG3D2021Chan}, among the widely used 3D generators, integrates NeRF into StyleGAN~\cite{karras2019style, karras2020analyzing, karras2021alias} by introducing tri-plane representation. Built on the success of EG3D, several approaches~\cite{sun2022ide, tang20233dfaceshop, Fruehstueck2023VIVE3D} use it as a prior for controllable 3D-aware facial image manipulation. In particular, Tang et al.~\cite{tang20233dfaceshop} incorporate guidance from a parametric head model into the generator to control illumination. However, the quality of their results is limited by the expressiveness of the parametric model.

\subsection{3D Relightable Face Generators}

In recent years, several approaches~\cite{pan2021shadegan, tan2022volux, FaceLit2023CVPR, Lumigan2023Deng} propose 3D relightable face generators by incorporating analytical shading models into their generators. For example, Pan et al.~\cite{pan2021shadegan} estimate shading using the Phong illumination model and multiplies it with the synthesized albedo to reconstruct the relit image. Deng et al.~\cite{Lumigan2023Deng} utilize a more complex shading model and handle visibility. The shading models used by these approaches, however, are not able to handle effects such as subsurface scattering which are necessary for faithful reproduction of the appearance of facial skin. To address this issue, Jiang et al.~\cite{NeRFFaceLighting2023Jiang} (\NFL{}) propose to implicitly model the shading by distilling EG3D into two separate triplane representations. Howerver, their approach has difficulty in maintaining the consistency of skin color in relit images and synthesizes albedo images that are biased towards lighter skin tones. The goal of our paper is to determine and address the underlying causes of these issues.

\subsection{Portrait Relighting}

\change{A large number of approaches~\cite{zhou2019dpr, sun2019relighting, nestmeyer2020relighting, pandey2021relighting, hou2021towards, hou2022face, yeh2022relighting, ponglertnapakorn2023difareli} perform relighting from a single 2D image. These approaches typically need to estimate the reflectance and geometry, and thus rely on either lab-captured or synthetic images for training. Unfortunately, such a reliance hampers the generalization capabilities of these algorithms. Furthermore, the 3D structure of the portrait is often not fully taken into consideration during the relighting, leading to suboptimal results, particularly in challenging lighting conditions.}

\subsection{Bias in Light/Albedo Estimation}

Estimating the image formation factors, particularly albedo and illumination, from a single image is a highly ambiguous task. A couple of recent techniques~\cite{Feng:TRUST:ECCV2022,Ren_2023_CVPR}, demonstrate that current techniques~\cite{aldrian2012inverse, hu2013facial, deng2019accurate, DECA2021Siggraph, egger2018occlusion, Feng:TRUST:ECCV2022, Ren_2023_CVPR} are biased towards estimating albedo images with lighter skin colors, and propose various ways to mitigate the problem. For example, Feng et al.~\cite{Feng:TRUST:ECCV2022} propose to use the full scene to disambiguate lighting and appearance. Despite producing promising results, we cannot utilize their method for unbiased light estimation as we only have access to portrait images. Ren et al.~\cite{Ren_2023_CVPR} utilize text to image models to force the estimated albedo and input images have similar skin tones. However, they only focus on albedo estimation and do not present a strategy for unbiased light estimation. 

Notably, the work by Legendre et al.~\cite{legendre2020learning} focuses on estimating high frequency lighting from portrait images, striving for enhanced precision and reduced bias in light estimation. To achieve this, they use light stage data of 70 subjects to train a light estimation network. Unfortunately, such a small scale dataset inherently lacks diversity in subjects, expressions, and accessories. Consequently, the performance of this approach on real portrait images with diverse characteristics could be suboptimal.




\section{Method}
\label{sec:method}

Built upon the pre-trained 3D face generator model by Chen et al.~\cite{EG3D2021Chan} (EG3D), \NFL{}~\cite{NeRFFaceLighting2023Jiang} distills the fused appearance and lighting information in the original triplane, into two triplanes. In this approach, one triplane encodes the geometry and albedo information, while the other is only responsible for producing the shading. Given a random latent vector $\vect{z}$, camera pose $\vect{v}$, and lighting condition $\vect{l}$, \NFL{} produces a relit image $G(\vect{z}, \vect{v}, \vect{l})$, through the product of a synthesized shading $S(\vect{z}, \vect{v}, \vect{l})$ and albedo $A(\vect{z}, \vect{v})$, i.e., $G(\vect{z}, \vect{v}, \vect{l}) = S(\vect{z}, \vect{v}, \vect{l}) A(\vect{z}, \vect{v})$. Note that only the shading is conditioned on the lighting as appearance (albedo) is independent of the illumination. 

Training the entire system is done using two discriminators: one exactly follows EG3D and is only conditioned on the camera pose, while the other is additionally conditioned on lighting to ensure the relit images are consistent with the lighting condition. This additional discriminator, necessitates extracting lighting information from the real images in the training data. \NFL{} does so using SfSNet~\cite{sfsnetSengupta18} which represents the light with nine $2^\text{nd}$ order spherical harmonics (SH) coefficients, $\vect{l} \in \mathbb{R}^9$.


As discussed, although the implicit lighting representation allows this approach to model intricate shading details, this flexibility introduces two major issues. First, when lighting condition comes from individuals with dark skin tones, this approach produces relit images with dark skin, regardless of the albedo's skin color. As shown in Fig.~\ref{fig:alg_2_analysis}, the shading image corresponding to the target lighting from a dark skin individual (Target 2), exhibits darkening of the eyes and bright highlights on the cheeks, forehead, and lips, resulting in a relit image with an altered skin color. Second, \NFL{} exhibits a significant bias towards albedo maps with lighter skin tones, as shown in Fig.~\ref{fig:motivation_separation}.

In generative adversarial networks (GANs), the generator attempts to follow the distribution of the training data. Since, \NFL{} uses two conditional discriminators, the generator is forced produce relit images that match the following two data distributions: $p_{\text{data}}(\vect{I}|\vect{l}, \vect{v})$ and $p_{\text{data}}(\vect{I}|\vect{v})$. Our key observation is that the generator produces dark skin tones for certain lightings in an attempt to match the training data distribution $p_{\text{data}}(\vect{I}|\vect{l}, \vect{v})$, i.e., the SH coefficients and skin tone for training images of individuals with dark skin are correlated. This correlation stems from the existence of bias in the estimated SH coefficients by SfSNet~\cite{sfsnetSengupta18}. 

Note that this also explains the reason behind \NFL{}'s bias towards albedos with lighter skin tones. The generator should produce relit images that follow the distribution $p_{\text{data}}(\vect{I}|\vect{v})$. That means the skin tone distribution of generated and the training images should be similar. Since the generator can produce images of individuals with dark skin tone through shading, it can match the distribution of the training data without resorting to producing albedo's with dark skin tone. Therefore, addressing the skin tone consistency of relit images will automatically fix the albedo's bias toward lighter skin colors.

In the following sections, we perform an analysis to better understand the bias and present our solution.


\subsection{Analysis}
\label{sec:analysis}


We begin by conducting an experiment to verify our observation that the problems originate from the bias in predicted SH coefficients. To do so, we randomly select 50,000 images from FFHQ~\cite{karras2019style} and classify them based on the skin tone. To classify the images, we utilize the CLIP model~\cite{clip-v139-radford21a} and evaluate the similarity between a set of four texts, describing the skin tone, and the input image. Specifically, we use the text ``a photo of a person with \{$c$\} skin tone'', where $c \in \{\text{fair, medium, tan, dark}\}$. We pick the text with the highest similarity as the skin tone of the individual.

\begin{figure}
\vspace{-0.1in}
\centering
\includegraphics[width=0.95\linewidth, scale=1.0, angle=-0]{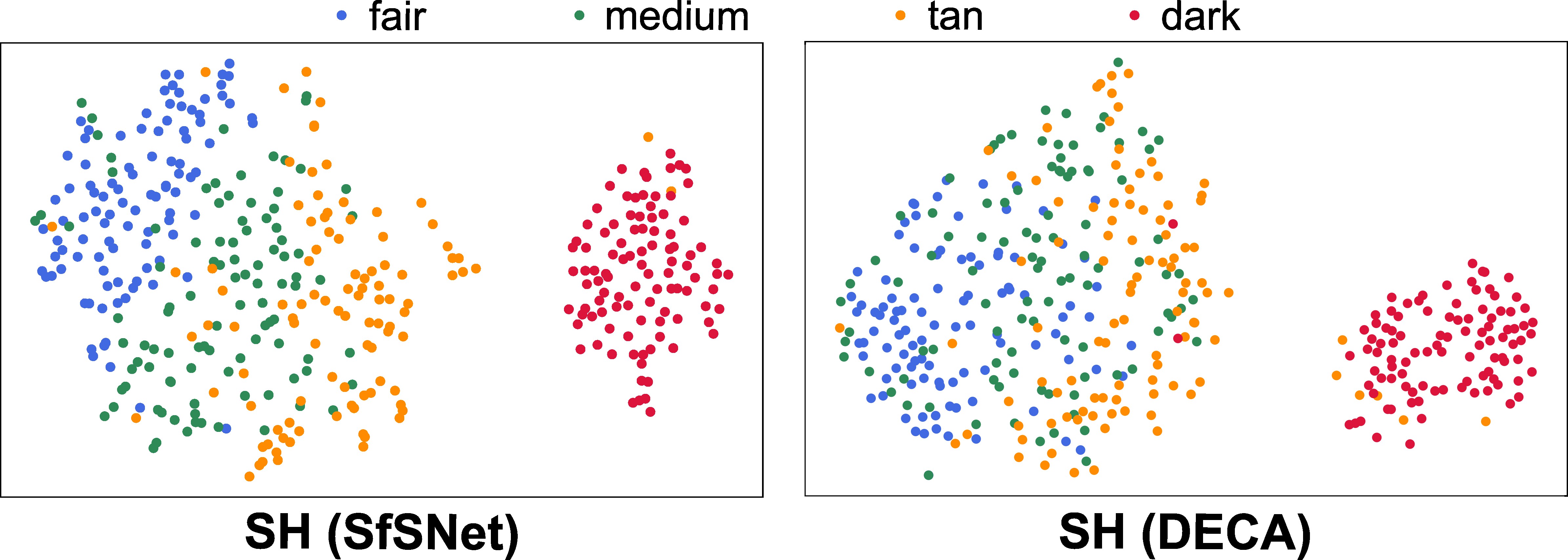}
\vspace{-0.1in}
\caption{We visualize the $2^\text{nd}$ order SH coefficients estimated using SfSNet~\cite{sfsnetSengupta18} and DECA~\cite{DECA2021Siggraph} from 400 images with different skin colors (100 in each category). We use t-SNE to visualize the coefficients in 2D. The coefficients extracted from images with dark skin form a distinct cluster in both cases.}
\vspace{-0.2in}
\label{fig:tsne_dpr_deca}
\end{figure}

Subsequently, we randomly sample 100 images from each class and extract their SH coefficients using SfSNet~\cite{sfsnetSengupta18}, following the methodology of \NFL{}~\cite{NeRFFaceLighting2023Jiang}. We then visualize these 400 9-dimensional SH coefficients in a two-dimensional space using t-SNE, which is a nonlinear dimensionality reduction technique. As depicted in Fig.~\ref{fig:tsne_dpr_deca} (left), the SH coefficients extracted from images with dark skin tones (shown in red) are distinctly clustered away from the other non-dark SH coefficients. Because of this bias, the discriminator can easily find significant correlation between the clustered SH coefficients and darker skin tones and force the generator to produce similar results. Note that this bias is not unique to SfSNet and exists in other portrait light estimation techniques as well. To demonstrate this, we perform the same analysis, but using the SH coefficients estimated by DECA~\cite{DECA2021Siggraph}. As shown in Fig.~\ref{fig:tsne_dpr_deca} (right), the SH coefficients from dark skin tones once again form a separate cluster.  

\begin{figure}
\vspace{-0.1in}
\centering
\includegraphics[width=0.95\linewidth, scale=1.0, angle=-0]{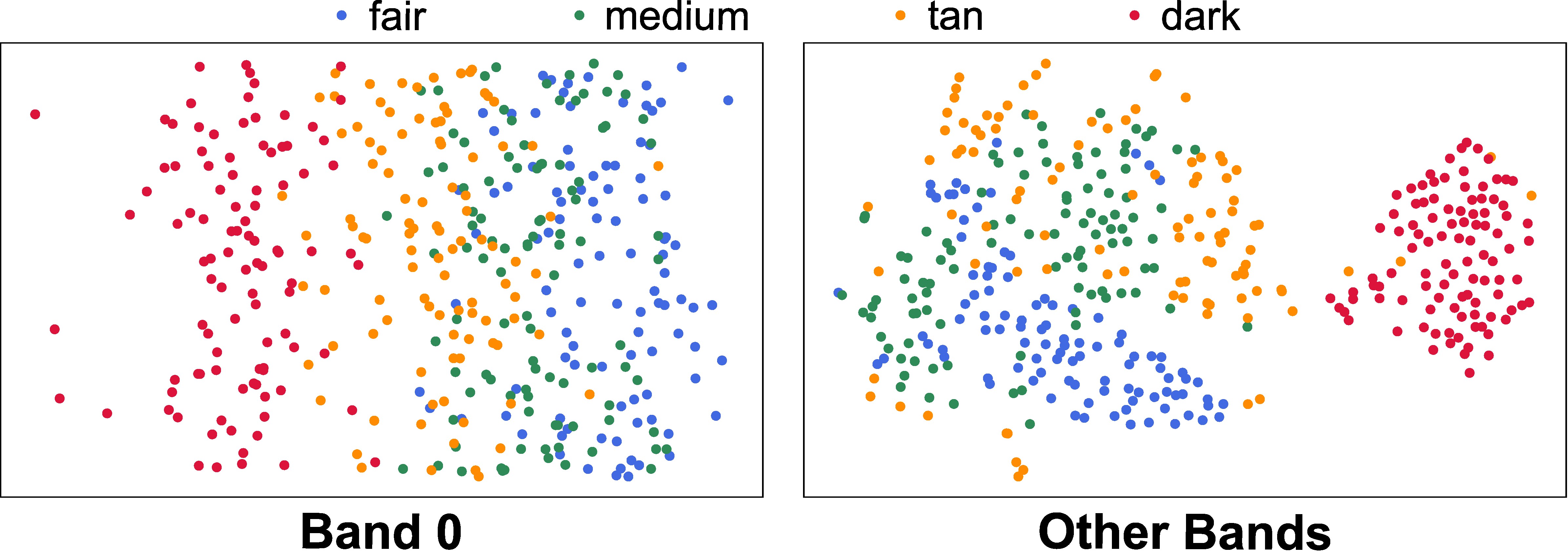}
\vspace{-0.1in}
\caption{\change{We visualize the SH coefficients, estimated by SfSNet, in band 0 and other bands. We augment the one dimensional coefficients in band 0 with an additional randomly filled dimension for better visualization. For other bands, however, we use t-SNE to reduce the dimensions from eight to two. As seen, the bias is not limited to the magnitude of the lighting (band 0) and appears in other higher order SH coefficients as well.}}
\vspace{-0.05in}
\label{fig:tsne_dpr_degree012}
\end{figure}

Given this observation, a question naturally arises: how do these SH coefficients differ from those extracted from non-dark faces? Estimating albedo and lighting from a single image is a highly ambiguous task with many plausible solutions. For example, an image can be explained by a dark albedo and bright light, as well as a light albedo and dim light. Recent studies~\cite{Feng:TRUST:ECCV2022,Ren_2023_CVPR} have shown that the current albedo/light estimation methods are biased towards producing albedos with lighter skin tones; they compensate this bias by estimating dimmer lighting. To verify this, we visualize the SH coefficients in band 0 (corresponding to the lighting magnitude) in Fig.~\ref{fig:tsne_dpr_degree012} (left). Note that to better visualize the one dimensional band 0 coefficients, we add an extra dimension, filled with random values between 0 and 1, and show the scatter plot of these 2D points. As seen in Fig.~\ref{fig:tsne_dpr_degree012} (left), the coefficients corresponding to dark skin tones have generally smaller DC terms (left is smaller values), and thus represent dimmer lightings.

We further visualize the SH coefficients in other bands to determine if the bias in band 0 fully accounts for the bias in the SH coefficients. Fig.~\ref{fig:tsne_dpr_degree012} (right) show this visualization where we use t-SNE to reduce the dimension of other bands from eight to two. Surprisingly, other bands show even stronger bias towards individuals with dark skin tones.  This demonstrates that the bias not only manifests itself in the magnitude of lighting, but is also encoded in the directions provided by higher order SH coefficients.

\begin{figure}
\centering
\vspace{-0.1in}
\includegraphics[width=0.95\linewidth, scale=1.0, angle=-0]{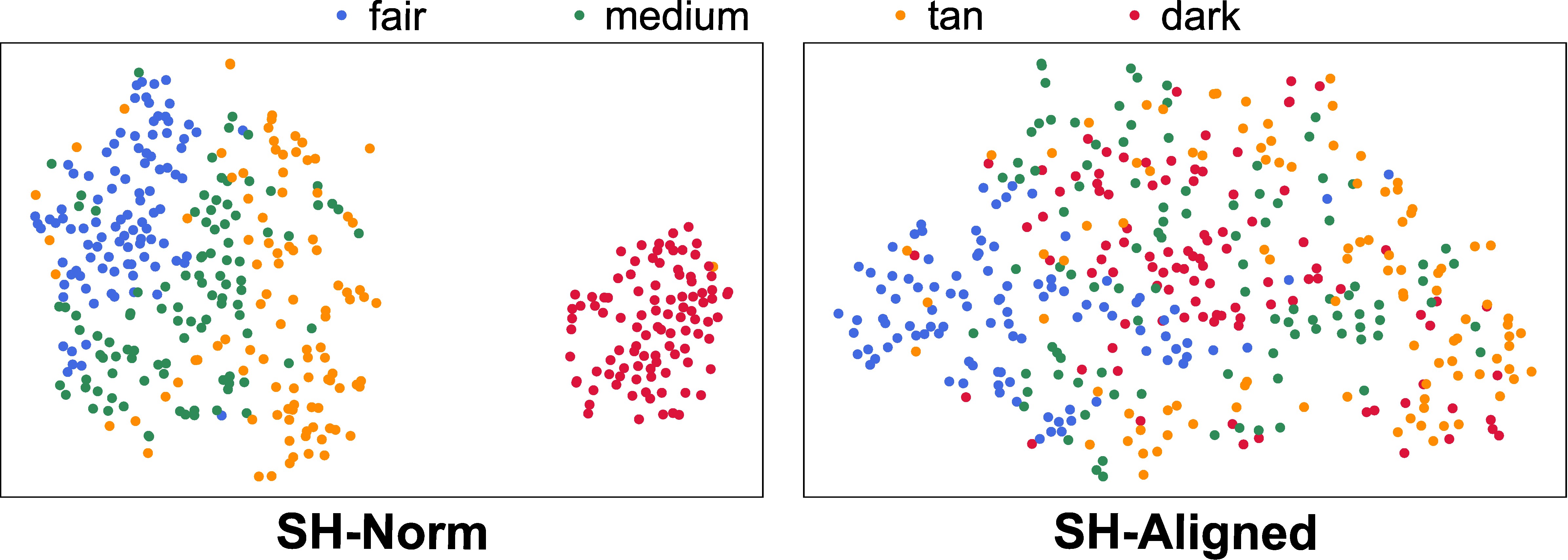}
\vspace{-0.1in}
\caption{\change{On the left, we visualize the SH coefficients after normalization. On the right, we showcase the normalized SH coefficients with statistical alignment of the coefficients of dark to non-dark skin tones. This approach effectively mitigates bias in the estimated SH coefficients.}}
\vspace{-0.25in}
\label{fig:tsne}
\end{figure}

\subsection{Mitigating Bias}
\label{ssec:mitigate_bias}

To address the problem of skin tone consistency and bias in relightable generators, we need to tackle the source of the issue, i.e., use unbiased SH coefficients during training. The obvious solution is to use an unbiased light estimation method. However, as demonstrated in Sec.~\ref{sec:analysis}, even the widely used technique by Feng et al.~\cite{DECA2021Siggraph}, estimates biased SH coefficients. Note that while there have been a couple of recent attempts to mitigate the skin tone bias~\cite{Feng:TRUST:ECCV2022,Ren_2023_CVPR,legendre2020learning}, these techniques either only focus on albedo estimation, require the full scene (not just portrait images), or could have difficulty generalizing to real images with diverse subjects, expression, and poses. 

Therefore, we opt for using the existing biased light estimation methods, but propose a simple strategy to mitigate their bias and prevent the discriminator from finding correlation between the coefficients and skin tone. Specifically, to combat the bias in band 0, we propose to normalize the SH coefficients by their DC term, i.e. $\vect{l}_n = \vect{l}[0:9]/\vect{l}[0]$. Normalizing the coefficients by the DC term completely removes the bias from the coefficients in band 0.

To address the bias in the other bands, we make a key observation that the illumination of the scene in the training data is independent of the individual's skin color. Therefore, the statistics (mean and standard deviation) of the SH coefficients of the images with dark skin colors, should be similar to the ones with non-dark skin tones. Armed by this observation, we propose to compute the mean and standard deviation of the SH coefficients corresponding to dark and non-dark individuals over the entire training data. We use the CLIP-based strategy presented in Sec.~\ref{sec:analysis} to place the images into dark and non-dark (fair, medium, and tan) categories. We then adjust the SH coefficients of images with dark skin tone to match the statistics of the non-dark individuals as follows:

\vspace{-0.15in}
\begin{equation}
    \vect{l}_{\text{nsa}}[i] = \frac{\vect{l}_n[i] - \mu_d[i]}{\sigma_d[i]}\sigma_{\text{nd}}[i] + \mu_{\text{nd}}[i],
\end{equation}
\vspace{-0.15in}

\noindent where $\vect{l}_{\text{nsa}}[i]$ is the $i^{\text{th}}$ normalized and statistically-aligned SH coefficient. Moreover, $\mu_d$, $\sigma_d$, $\mu_{\text{nd}}$, and $\sigma_{\text{nd}}$ are the mean and standard deviation of the normalized SH coefficient of dark and non-dark images, respectively. Note that for images with non-dark individuals we use the normalized SH coefficients without any other modifications, i.e., $\vect{l}_{\text{nsa}} = \vect{l}_n$.

\change{As shown in Fig.~\ref{fig:tsne}, the $2^\text{nd}$ order coefficients after normalization still display noticeable clusters, whereas the normalized coefficients after statistical-alignment show no discernible clusters, indicating effective bias mitigation.}

Special care should be taken when using the normalized SH coefficients during training. To minimize the GAN loss, the generated relit images (input to the discriminator) should match the distribution of training data with diverse illumination magnitude. On the other hand, our generator takes normalized (and statistically-aligned) SH coefficients as the input and is expected to produce relit images with constant illumination magnitude. Training the system without considering this mismatch will force the generator to produce images with diverse illumination magnitude, even with normalized SH coefficients as the input.

To address this issue, we propose to randomly adjust the magnitude of the illumination of the generated images during training. Because of the linearity of the light, adjusting the magnitude of the illumination can be done in the image domain as follows:

\vspace{-0.15in}
\begin{equation}
\label{eq:LinearScale}
    \hat{I} = \left(G(\vect{z}, \vect{v}, \vect{l}_{\text{nsa}})^{\gamma} \times s \right)^{\frac{1}{\gamma}}
\end{equation}
\vspace{-0.15in}

\noindent where $\gamma = 2.2$ in our implementation and $s$ is the scaling factor. Here, we first take the generated image with normalized illumination into the linear domain with gamma expansion. We then transform the scaled image into tonemapped domain with gamma compression.

The key idea is to force the generator to produce relit images with an illumination magnitude equal to the average light intensity over the training data by carefully setting the distribution of the scaling factor $s$. This can be achieved by calculating $s$ as follows:

\vspace{-0.15in}
\begin{equation}
    s = \frac{m(I_i)}{\frac{1}{\vert \mathcal{N} \vert}\sum_{n \in \mathcal{N}} m(I_n)},
\end{equation}
\vspace{-0.15in}

\noindent where $I_i$ is a randomly selected training image, $m(I_i)$ is the illumination magnitude of the image, and $\mathcal{N}$ refers to the set of all training images. Since $s$ is computed based on the ratio of magnitudes, we propose to approximate the illumination magnitude using average of the pixel intensities in the facial area. The only remaining caveat is that the ratio should be computed on images with similar skin tones. Our final formulation is thus as follows:

\vspace{-0.15in}
\begin{equation}
    s = \frac{m(I_i)}{\frac{1}{\vert \mathcal{N_c} \vert}\sum_{n \in \mathcal{N}_c} m(I_n)}, \ \text{where} \ m(I_i) = \frac{1}{\vert \mathcal{P} \vert}\sum_{p \in \mathcal{P}} I_i[p].
\end{equation}
\vspace{-0.15in}

Here, $\mathcal{N}_c$ refers to the subset of training images with the same skin tone (fair, medium, tan, and dark) as $I_i$. Moreover, $\mathcal{P}$ is a subset of pixels on the face. We demonstrate the impact of this magnitude scaling scheme in Sec.~\ref{ssec:magnitude_scaling}.

Once trained, our generator can produce relit images with constant illumination. To produce images with varying lighting magnitude, we simply use Eq.~\ref{eq:LinearScale} to adjust the illumination scale.

\section{Results}
\label{sec:results}

Throughout this section we analyze the performance of different versions of \NFL{} and compare them against our proposed strategy. Specifically, these variations include:
\vspace{-0.1in}

\begin{itemize}[itemsep=0.1pt]
    \item \textbf{\NFL{}}: The officially released checkpoint, provided by the authors.
    \item \textbf{\NFL{}-DECA}: Using SH coefficients estimated by DECA~\cite{DECA2021Siggraph}.
    \item \textbf{SH-Norm}: Using normalized SH coefficients.
    \item \textbf{Ours}: \change{Using normalized and statictically-aligned SH coefficients.}
\end{itemize}

We train all the variations on FFHQ~\cite{karras2019style} for 5M images, following the training strategy of \NFL{}. We use a batch size of 28 and perform the training on 4 A100 GPUs. Note that, in all cases except \textbf{\NFL{}-DECA}, we use SfSNet~\cite{sfsnetSengupta18} to obtain the SH coefficients.

\begin{figure}[t]
\vspace{-0.1in}
\centering
\includegraphics[width=0.86\linewidth, scale=1.0, angle=-0]{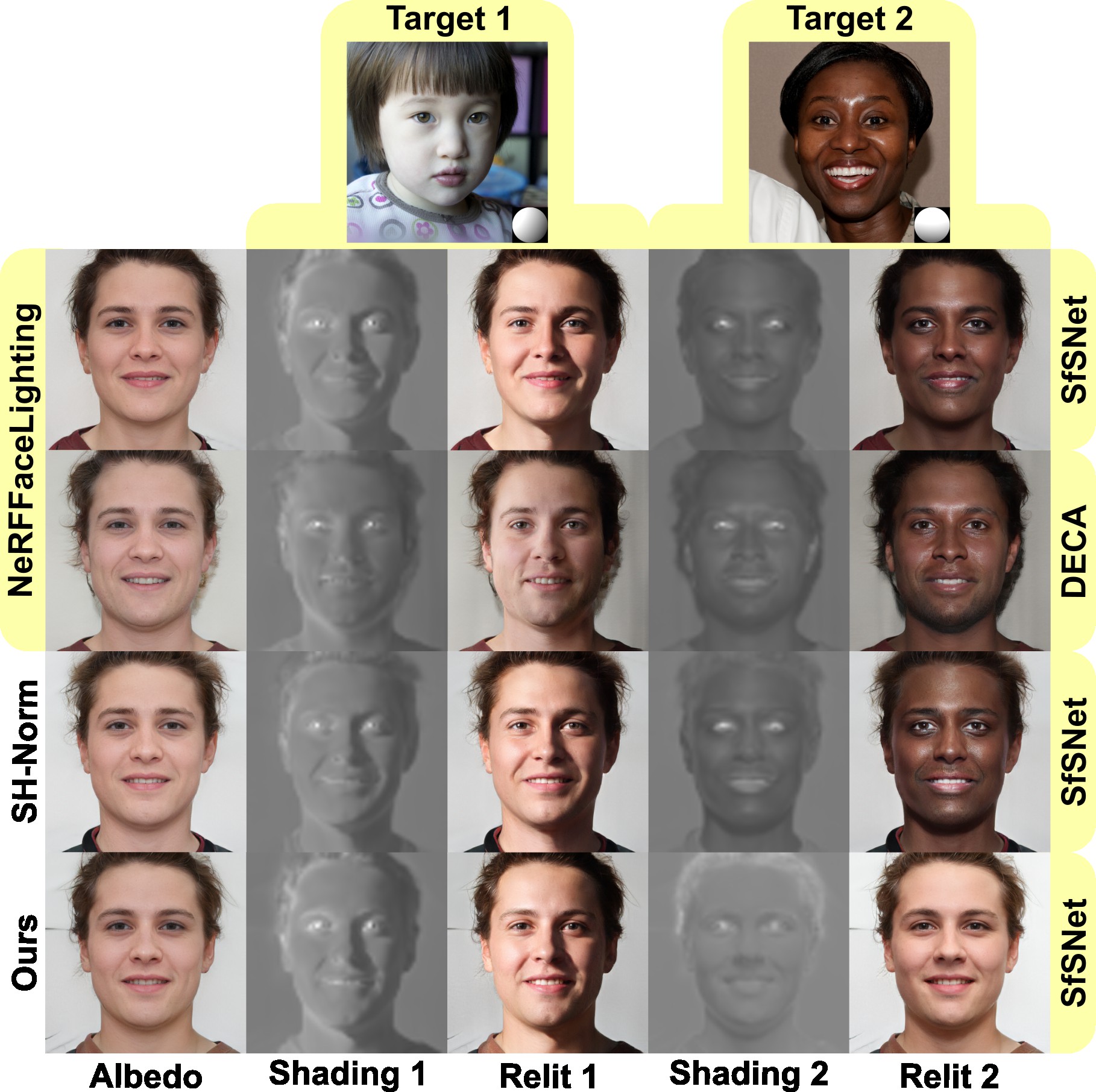}
\vspace{-0.1in}
\caption{We compare different approaches by producing relit images using two target lightings. Our approach produces results with consistent skin tone for both lightings.}
\vspace{-0.2in}
\label{fig:alg_2_analysis}
\end{figure}

\subsection{Relighting Sampled Images}
\label{subsec:skintoneconsistency}

In Fig.~\ref{fig:alg_2_analysis}, we compare how different variations preserve the skin tone in relit images. Specifically, we perform relighting using the SH coefficients extracted from two target images with fair (Target 1) and dark (Target 2) skin tones. As seen, while the sampled albedo has a fair skin tone, \textbf{\NFL{}} produces a relit image with dark skin for the second target. Similarly, \textbf{\NFL{}-DECA} is not able to produce relit images that are consistent with albedo's skin color. This demonstrates that both light estimation techniques (DECA and SfSNet) produce biased SH coefficients. Moreover, normalizing the SH coefficients (\textbf{SH-Norm}) alone is not sufficient. On the other hand, our approach using normalized and statistically-aligned SH coefficients produces relit images that have fair skin color and are consistent with the albedo. It is worth noting that, although the second target exhibits strong specular highlights, our approach correctly produces a relit image with a more diffuse appearance. This is because the effect of subsurface scattering is stronger in fair skin tones, and thus fair skins under the same illumination have a more diffuse appearance.

We further numerically evaluate the effectiveness of different variations in maintaining skin color consistency after relighting. Specifically, we randomly sample 100 latent codes corresponding to albedo maps with a fair skin color. We then relight each albedo with 100 randomly sampled lightings and measure the skin tone consistency between each relit image and its corresponding albedo.  To do so, we first use the CLIP model~\cite{clip-v139-radford21a} and follow the process described in Sec.~\ref{sec:analysis} to obtain a set of 4 similarity scores between each image and four skin tone categories (fair, medium, tan, dark). We then compute the cosine similarity between the scores for the relit image and its corresponding albedo as the measure of skin tone consistency. A value close to one corresponds to perfect skin tone consistency. Table~\ref{tab:relighting_consistency} shows the average (Avg.), standard deviation (STD), and minimum scores. Our method produces results with higher average and lower standard deviation, indicating superior performance in preserving facial color in relit images. Notably, our worst case scenario, indicated by the minimum score, is significantly better than \NFL{}.

\begin{table}[t] 
\vspace{-0.1in}
\caption{Quantitative comparison against the other approaches in terms of skin tone consistency metric.}
\vspace{-0.1in}
\centering
\begin{adjustbox}{width=0.42\textwidth,center}
\begin{tabular}{ lccc} 
\hline
\hline
& Avg. $\uparrow$ & STD $\downarrow$ & Minimum $\uparrow$ \\
\hline  
NeRFFaceLighting & 0.9704 & 0.0386 & 0.4404 \\
NeRFFaceLighting-DECA & 0.9622 & 0.0591 & 0.1065 \\
SH-Norm & 0.9652 & 0.0389 & 0.5912 \\
Ours & \textbf{0.9745} & \textbf{0.0221} & \textbf{0.6388} \\

\hline  
\end{tabular}
\end{adjustbox}
\label{tab:relighting_consistency}
\vspace{-0.1in}
\end{table}


\begin{figure}[t]
\centering
\includegraphics[width=0.86\linewidth, scale=1.0, angle=-0]{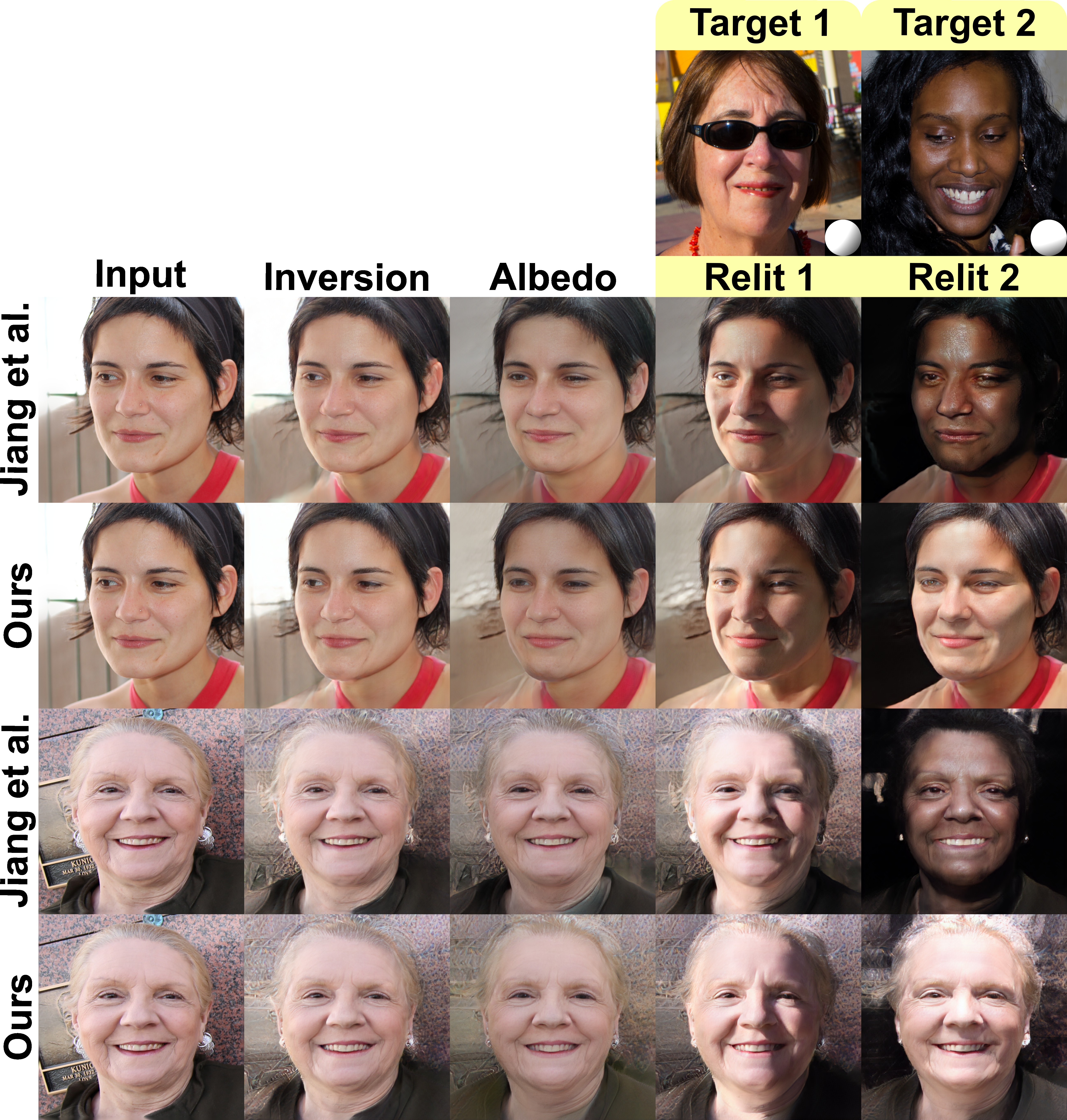}
\vspace{-0.1in}
\caption{Comparison against \NFL{ [Jiang et al.]} on real images. Our relit images have consistent skin colors, while \NFL{} produces results with significantly darker skin for the second target.}
\vspace{-0.2in}
\label{fig:real_image_relighting}
\end{figure}

\subsection{Relighting Real Images} 

We compare the ability of different variations in preserving the consistency of skin tone when relighting real images. To do so, we follow \NFL{}~\cite{NeRFFaceLighting2023Jiang} and project the input images into the latent space of the generator. We then relight the inverted images using lighting estimated from two target images with fair (Target 1) and dark (Target 2) skin tones. As shown in Fig.~\ref{fig:real_image_relighting}, \NFL{} produces relit images with different skin tones, while our relit images have consistent skin colors.

\begin{table}[t] 
\vspace{-0.1in}
\caption{KL divergence between the distribution of the skin colors of generated albedo images by different approaches and EG3D.}
\vspace{-0.1in}
\begin{adjustbox}{width=0.33\textwidth,center}
\centering
\begin{tabular}{ lc} 
\hline
\hline
&  KL divergence $\downarrow$ \\
\hline  
NeRFFaceLighting  & 0.0043\\
NeRFFaceLighting-DECA  & 0.0047\\
SH-Norm  & 0.0040\\ 
Ours  & \textbf{0.0029}\\

\hline  
\end{tabular}
\end{adjustbox}
\label{tab:albedo_distribution}
\vspace{-0.1in}
\end{table}


\subsection{Albedo's Skin Tone Bias}

\begin{figure}
\centering
\includegraphics[width=0.9\linewidth, scale=1.0, angle=-0]{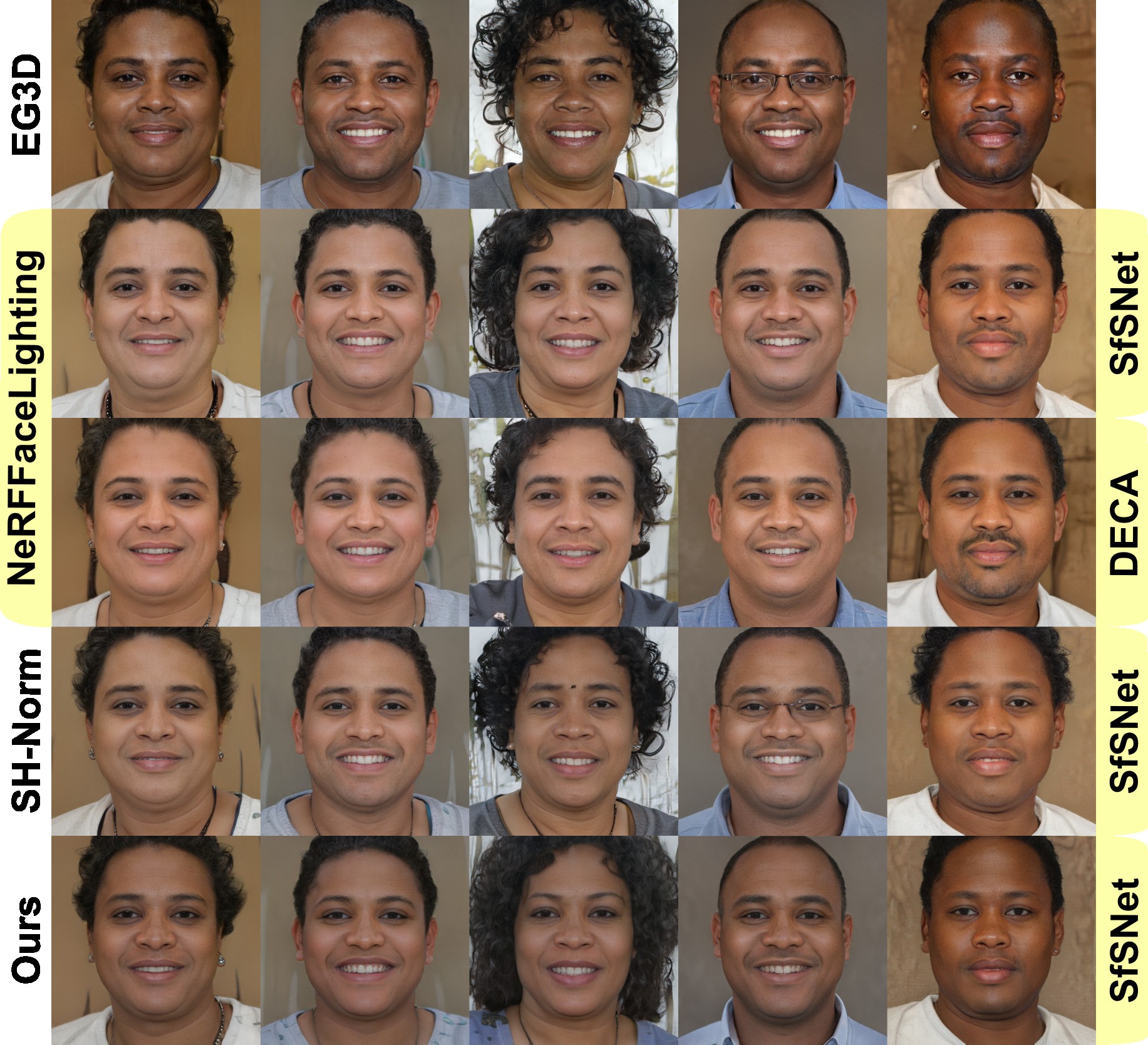}
\vspace{-0.1in}
\caption{We show several randomly sampled images from EG3D and the corresponding albedo images produced by different variations of \NFL{} and ours. When distilling EG3D, both versions of \NFL{} tend to produce albedos with lighter skin colors. On the other hand, our solution produces albedo images that better represents the skin color of the original EG3D samples. }
\vspace{-0.2in}
\label{fig:motivation_separation}
\end{figure}

We begin by showing a few synthesized albedo images with dark skin in Fig.~\ref{fig:motivation_separation}. When distilling the EG3D triaplane, \NFL{} trained with SH coefficients from both SfSNet and DECA has bias towards producing albedos with lighter skin colors, as evident by the results. Our solution, however, mitigates this bias and produces albedo images with darker skin tones, resembling the corresponding samples from EG3D. \change{Note that although the images in each column are produced by sampling the same latent vector, there are small variations in appearance. These variations, however, are expected as the models are trained independently. While we start with the EG3D generator in each case, fine-tuning the model using the adversarial loss leads to small variations in different attributes.}

Next, we numerically compare the ability of different methods in properly distilling the appearance (albedo) from EG3D. To do so, we randomly sample 50,000 latent codes and generate the corresponding albedo for each method. We then use the CLIP-based method, described in Sec.~\ref{sec:analysis}, to assign one of the four skin color categories to each albedo image. Following this, we obtain the distribution of generated albedos for each method in terms of the four categories. We also perform the same process and obtain the distribution of skin colors for EG3D. We then find the KL divergence between the skin color distribution of each method and EG3D. Smaller KL divergence means the approach is not introducing skin color bias on top of EG3D. As shown in Table~\ref{tab:albedo_distribution}, our results have the smallest KL divergence, demonstrating the effectiveness of our solution in mitigating the bias.

\subsection{Effect of Magnitude Scaling}
\label{ssec:magnitude_scaling}

\begin{figure}
\centering
\includegraphics[width=0.9\linewidth, scale=1.0, angle=-0]{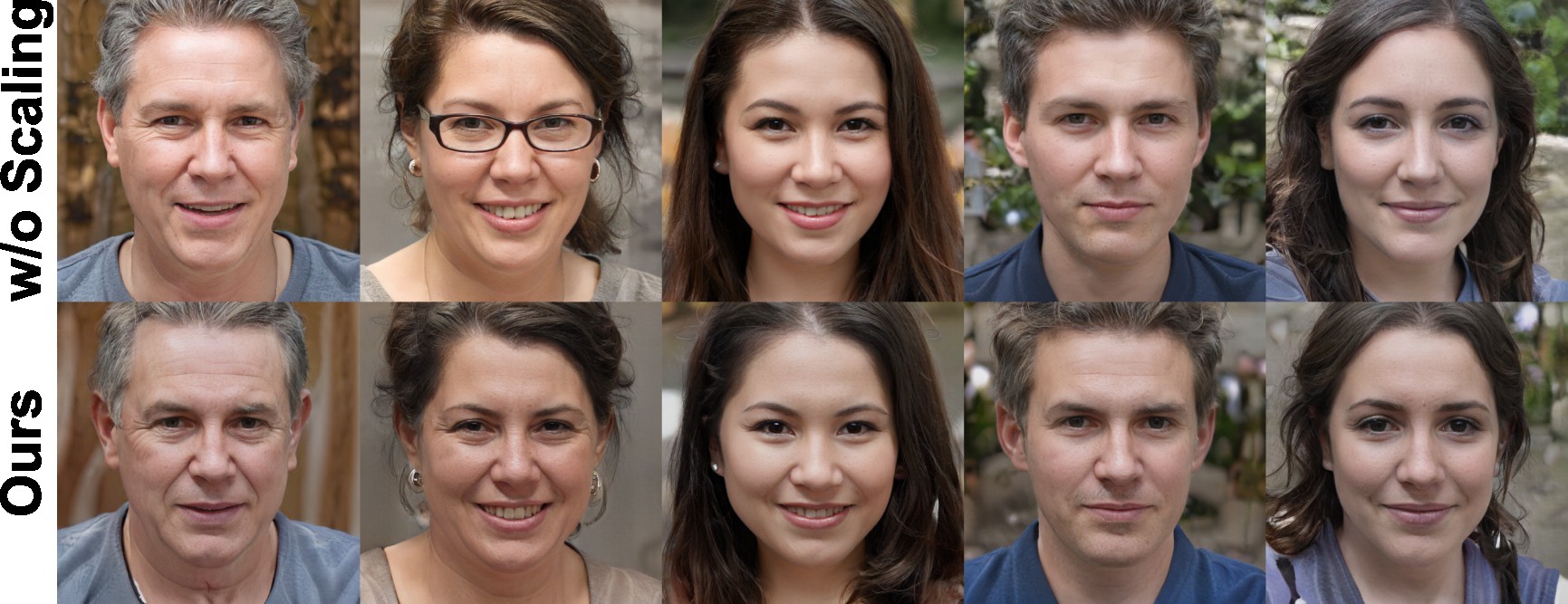}
\vspace{-0.1in}
\caption{\change{Evaluation of the impact of magnitude scaling scheme during training. When randomly sampling images with the same lighting condition, our results demonstrate constant illumination magnitude. In contrast, the generator trained without the scaling technique displays varying illumination magnitudes.}}
\vspace{-0.1in}
\label{fig:scaling}
\end{figure}

Here, we investigate the impact of the magnitude scaling scheme, discussed in Sec.~\ref{ssec:mitigate_bias} (Eq.~\ref{eq:LinearScale}), by comparing our approach against a variant of our technique that is trained without scaling. In Fig.~~\ref{fig:scaling}, we show images obtained by randomly sampling the latent code, but with fixed camera and lighting condition. As seen, the approach without scaling produces images with slight variation in the lighting magnitude, while our method exhibits constant illumination. 


We further evaluate the impact of this component numerically. We do so by first generating 1,000 relit images by randomly sampling the latent code, camera, and lighting (normalized SH coefficients). We then use DECA~\cite{DECA2021Siggraph} to estimate the illumination magnitude (DC of the SH coefficients) and report their standard deviation in Table~\ref{tab:dc_std}. Note that, to avoid DECA's bias, we ensure all the 1,000 generates images are of individuals with fair skin tones using the CLIP-based method, described in Sec.~\ref{sec:analysis}. As seen, our generator trained with scaling exhibits smaller magnitude variation.


\begin{table} 
\small
\caption{\change{Illumination Magnitude Consistency.}}
\vspace{-0.1in}
\begin{adjustbox}{width=0.28\textwidth,center}
\centering
\begin{tabular}{ lc} 
\hline
\hline
&  Standard deviation $\downarrow$ \\
\hline  
w/o Scaling  & 0.2349\\
Ours  & \textbf{0.1011}\\

\hline  
\end{tabular}
\end{adjustbox}
\label{tab:dc_std}
\vspace{-0.2in}
\end{table}

\subsection{Direct Applications}
\label{sec:discussion}

Our approach statistically mitigates the bias in SH coefficients, making it applicable for improving skin tone consistency in any relighting technique that uses SH coefficients as a condition. To demonstrate this, we apply our approach to another GAN-based 3D relightable generator, FaceLit~\cite{FaceLit2023CVPR}, and a recent diffusion-based relighting method, DiFaReli~\cite{ponglertnapakorn2023difareli}. Both of these methods use SH coefficients of lighting extracted via DECA~\cite{DECA2021Siggraph}. Note that here, we directly apply our approach to FaceLit~\cite{FaceLit2023CVPR} and DiFaReli~\cite{ponglertnapakorn2023difareli} without fine-tuning. 


\begin{figure}
\centering
\includegraphics[width=0.85\linewidth, scale=1.0, angle=-0]{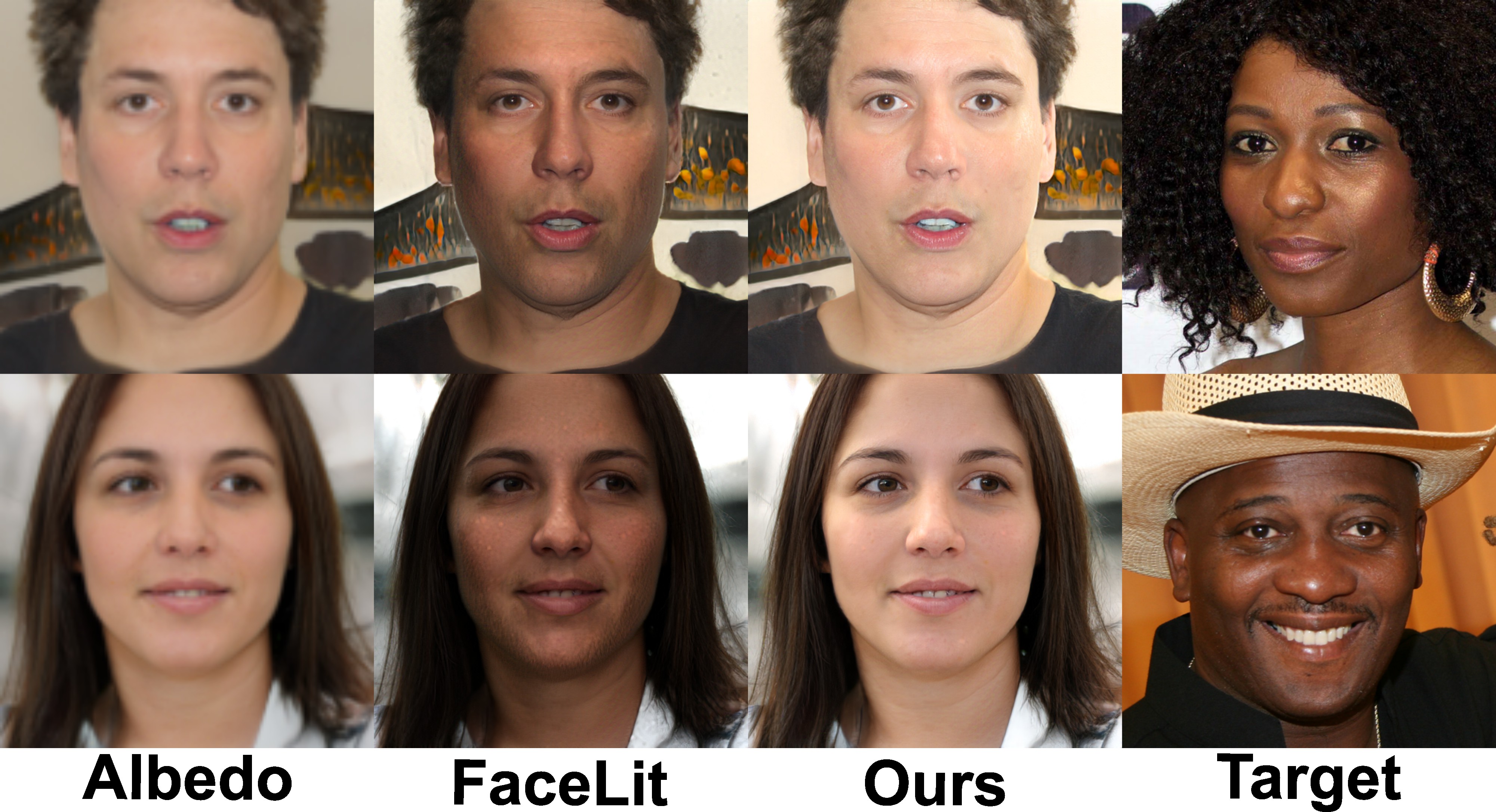}
\vspace{-0.1in}
\caption{Comparisons against FaceLit~\cite{FaceLit2023CVPR} on sampled albedo images.}
\label{fig:discussion_facelit}
\end{figure}

\begin{table}[t] 
\vspace{-0.1in}
\caption{Quantitative comparison against FaceLit~\cite{FaceLit2023CVPR} in terms of the consistency of skin tone and identity.}
\vspace{-0.0in}
\begin{adjustbox}{width=0.46\textwidth,center}

\centering
\begin{tabular}{ lcccccc} 
\hline
\hline
&  \multicolumn{3}{c}{Skin tone consistency} &  \multicolumn{3}{c}{Identity similarity} \\
\cmidrule(lr){2-4}
\cmidrule(lr){5-7}
& Avg. $\uparrow$ & STD $\downarrow$ & Minimum $\uparrow$ & Avg. $\uparrow$ & STD $\downarrow$ & Minimum $\uparrow$ \\ 
\cmidrule(lr){1-4}
\cmidrule(lr){5-7}
FaceLit & 0.9651 & 0.0439 & 0.7886 & 0.7542 & 0.1007 & 0.4716  \\
Ours & \textbf{0.9708} & \textbf{0.0381} & \textbf{0.8052} & \textbf{0.7908} & \textbf{0.0967} & \textbf{0.5272} \\

\hline

\end{tabular}
\end{adjustbox}
\label{tab:relighting_consistency_facelit}
\vspace{-0.0in}
\end{table}


FaceLit~\cite{FaceLit2023CVPR}, similar to \NFL{}, produces relightable 3D portraits by incorporating lighting condition into EG3D~\cite{EG3D2021Chan}. We follow their recommended approach to generate albedo images by rendering them before the superresolution module under constant illumination. We then relight the albedo images using SH coefficients extracted via DECA~\cite{DECA2021Siggraph}. As shown in Fig.~\ref{fig:discussion_facelit}, with our statistically aligned SH coefficeints, the relit images exhibit improved skin tone consistency.
To further numerically compare our approach with FaceLit~\cite{FaceLit2023CVPR}, we randomly sample 100 albedo images and relight each with 100 random lighting conditions from the FFHQ dataset. 
We then compute skin tone consistency score, as introduced in Sec.~\ref{subsec:skintoneconsistency}, and identity similarity score using ArcFace\cite{deng2019arcface}, a face recognition method, between the sampled albedo and relit images. 
As shown in Table~\ref{tab:relighting_consistency_facelit}, our method performs better across all metrics, with higher averages, lower standard deviations, and larger minimum values in terms of both skin tone consistency and identity similarity scores.

\begin{figure}
\centering
\includegraphics[width=0.85\linewidth, scale=1.0, angle=-0]{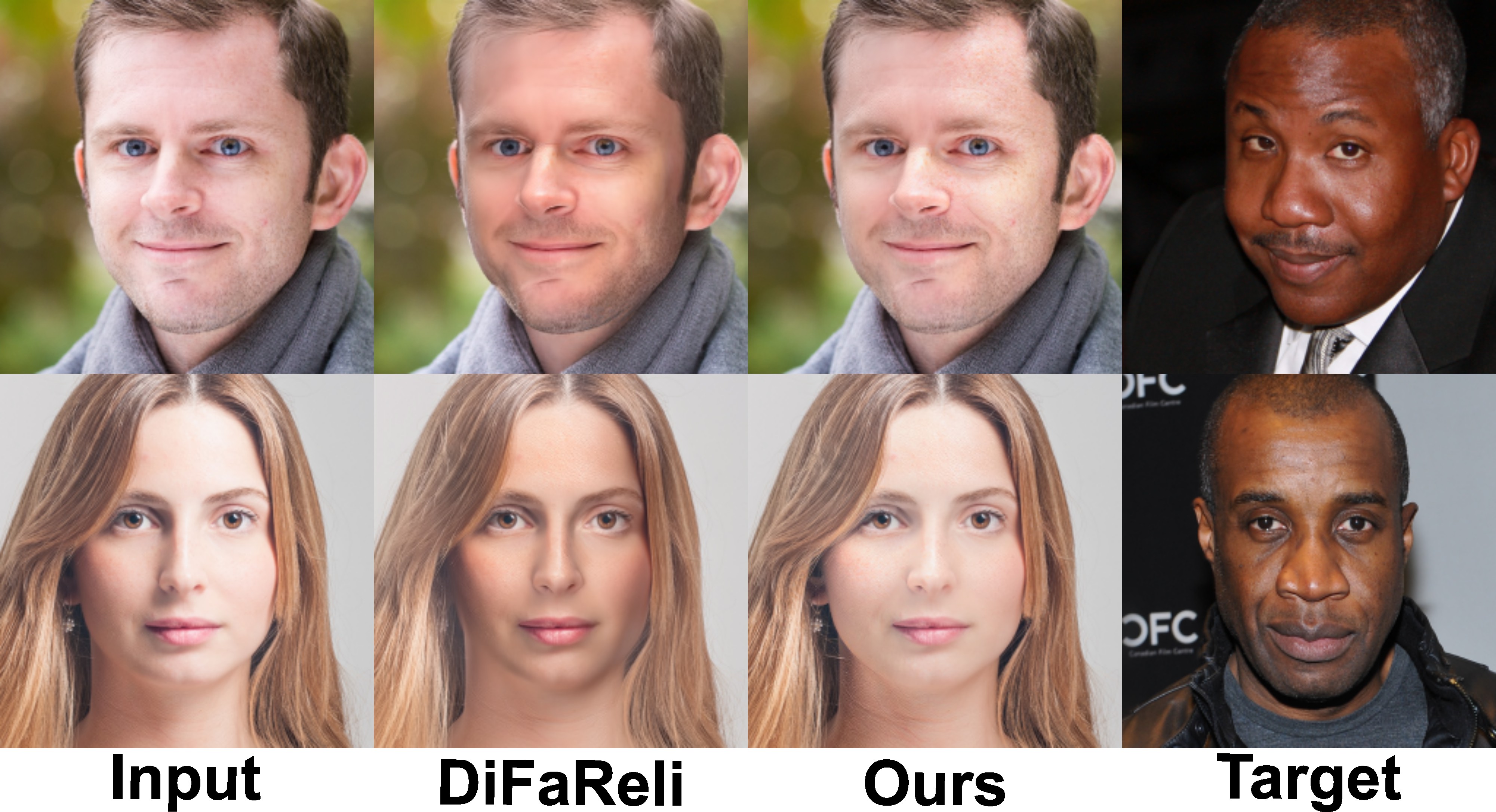}
\vspace{-0.1in}
\caption{Comparisons against DiFaReli~\cite{ponglertnapakorn2023difareli} on real images.}
\vspace{-0.2in}
\label{fig:discussion_difareli}
\end{figure}

\begin{table}[t] 
\vspace{-0.1in}
\caption{Quantitative comparison against DiFaReli~\cite{ponglertnapakorn2023difareli} in terms of the consistency of skin tone and identity.}
\vspace{-0.1in}
\begin{adjustbox}{width=0.46\textwidth,center}

\centering
\begin{tabular}{ lcccccc} 
\hline
\hline
&  \multicolumn{3}{c}{Skin tone consistency} &  \multicolumn{3}{c}{Identity similarity} \\
\cmidrule(lr){2-4}
\cmidrule(lr){5-7}
& Avg. $\uparrow$ & STD $\downarrow$ & Minimum $\uparrow$ & Avg. $\uparrow$ & STD $\downarrow$ & Minimum $\uparrow$ \\ 
\cmidrule(lr){1-4}
\cmidrule(lr){5-7}
DiFaReli & 0.9846 & 0.0257 & 0.6973 & 0.7622 & 0.1568 & 0.1913  \\
Ours & \textbf{0.9894} & \textbf{0.0205} & \textbf{0.8096} & \textbf{0.8170} & \textbf{0.1280} & \textbf{0.3332} \\

\hline  
\end{tabular}
\end{adjustbox}
\label{tab:relighting_consistency_difareli}
\vspace{-0.2in}
\end{table}


We further visually compare our approach against DiFaReli~\cite{ponglertnapakorn2023difareli}, a relighting method that utilizes a conditional diffusion model, in Fig.~\ref{fig:discussion_difareli}. As seen, this technique produces results with inconsistent skin tone using the SH coefficients estimated by DECA~\cite{DECA2021Siggraph} from the target images with dark skin tones. However, using our normalized and statistically aligned SH coefficients, it is able to produce results with better skin tone consistency.
We further numerically demonstrate the effectiveness of our approach in preserving skin tone consistency and identity similarity in Table~\ref{tab:relighting_consistency_difareli}. As seen, our method produces better results with better skin tone consistency and identity similarity, compared to DiFaReli with the original DECA coefficients.

\section{Conclusion, Limitations, and Future Work}

We have presented a comprehensive analysis and solution to the problem of the skin tone inconsistency and bias in the relightable face generator with implicit lighting representation. We observe that the issue stems from the bias in estimated lighting, which presents itself not only in the magnitude of illumination (band 0), but also in the other higher order bands of the spherical harmonics coefficients. Based on this observation we suggest performing the training using normalized and statistically-aligned SH coefficients. We demonstrate that this simple solution is highly effective in mitigating the bias and preserving the skin tone consistency of relit images, \notes{produced by a few GAN-based and diffusion-based relighting methods, highlighting the broad applicability of our method}.

Our method relies on proper estimation of the skin tone category using the CLIP model. While we demonstrate that our solution is highly effective, there may be cases where the CLIP model incorrectly detects the skin category, resulting in an incorrect alignment. We believe that mitigating this issue by investigating an unbiased light estimation method would be an interesting future research direction.

 Moreover, although our solution significantly reduces the bias, and improves the quality compared to \NFL{}, there is still a small gap in image quality between our approach and the backbone EG3D (see Table \notes{1 of the supplementary document}). In the future, we would like to explore potential approaches to further reduce this gap.

\section{Acknowledgements}

We sincerely thank the anonymous reviewers for their valuable feedback and constructive suggestions. Additionally, portions of this research were conducted with the advanced computing resources provided by Texas A\&M High Performance Research Computing.

\newpage

{\small
\bibliographystyle{ieee_fullname}
\bibliography{egbib}
}

\end{document}